\pdfoutput=1

\documentclass[11pt]{article}
\usepackage[unicode]{hyperref}

\usepackage{acl}

\usepackage{times}
\usepackage{latexsym}
\usepackage{amsmath}
\usepackage{amssymb}
\usepackage[pdftex]{graphicx}

\usepackage{epstopdf}

\usepackage[T2A,T1]{fontenc}
\usepackage{ucs}
\usepackage[utf8x]{inputenc}
\usepackage[belarusian,greek,thaicjk,vietnamese,english]{babel}
\addto\extrasthaicjk{\fontencoding{C90}\selectfont}
\makeatletter
\@namedef{opt@inputenc.sty}{utf8}
\makeatother
\usepackage{CJKutf8}

\usepackage{microtype}
\usepackage{booktabs}
\usepackage{comment}
\usepackage{multirow}

\makeatletter
\newcommand\footnoteref[1]{\protected@xdef\@thefnmark{\ref{#1}}\@footnotemark}
\makeatother

\usepackage{inconsolata}
\usepackage{graphicx}

\title{Tracing the Roots of Facts in Multilingual Language Models:\\
Independent, Shared, and Transferred Knowledge}

\author{Xin Zhao \\
 The University of Tokyo \\
 \texttt{xzhao@tkl.iis.u-tokyo.ac.jp} \\\And
 Naoki Yoshinaga\qquad\quad Daisuke Oba \\
 Institute of Industrial Science,\\
 The University of Tokyo\\
 \texttt{\{ynaga,oba\}@iis.u-tokyo.ac.jp}\\}

\begin{document}

\newcommand{\yn}[1]{\textcolor{black}{#1}}
\newcommand{\zx}[1]{\textcolor{black}{#1}}
\newcommand{\cshin}[1]{\textcolor{black}{#1}}
\newcommand{\nc}[1]{\textcolor{black}{#1}}

\maketitle
\begin{abstract}
Acquiring factual knowledge \yn{for language models (LMs)} in low-resource languages poses a 
\nc{serious} challenge\yn{,} 
\yn{thus resorting to}
cross-lingual transfer in \yn{multilingual LMs (ML-LMs)}.
\cshin{In this study, we ask how \yn{ML-LMs acquire and represent
factual knowledge.}}
\yn{\nc{Using} the multilingual factual knowledge probing dataset, mLAMA,
we first conducted a neuron
\nc{investigation} \cshin{of ML-LMs (\nc{specifically}, multilingual BERT).}}
We then traced the roots of facts \yn{back} to 
the knowledge source (Wikipedia)
to identify the 
ways in which ML-LMs \nc{acquire} specific facts.
\yn{We finally}
\nc{identified} 
three 
patterns \yn{of} \nc{acquiring} and representing 
facts in ML-LMs: language-independent, cross-lingual shared and transferred, and \nc{devised} methods 
\nc{for differentiating}
them.
\cshin{Our findings highlight the challenge of maintaining 
consistent 
factual knowledge across languages, \yn{underscoring} the need for 
better 
fact representation learning \yn{in ML-LMs}}.\footnote{
\yn{code:} \url{https://github.com/xzhao-tkl/fact-cl}}

\end{abstract}

\section{Introduction}
To mitigate the inherent data sparseness \nc{of} low-resource languages,
multi-lingual language models (ML-LMs) such as 
\cshin{mBERT~\cite{devlin-etal-2019-bert}, XLM-R~\cite{conneau-etal-2020-unsupervised}, mT5~\cite{xue-etal-2021-mt5}, and BLOOM~\cite{workshop2023bloom} 
have been developed}
to transfer knowledge across languages.
The effectiveness of this cross-lingual transfer \nc{in ML-LMs has been demonstrated} on various 
language tasks~\cite{wu-dredze-2019-beto, chi-etal-2020-finding, pires-etal-2019-multilingual, huang-etal-2023-languages}. 
\nc{However}, a more challenging task is the cross-lingual transfer of specific factual 
knowledge, such as
\cshin{``Greggs is a British bakery chain.''}
In many low-resource languages, text data about such \cshin{knowledge} might be minimal or non-existent. 
Effectively transferring knowledge is \cshin{vital} for applications that 
handle factual knowledge, such as fact checking and relation extraction~\cite{lee-etal-2020-language, verlinden-etal-2021-injecting}.

\begin{figure}[t]
    \centering
    \includegraphics[width=\linewidth,clip]{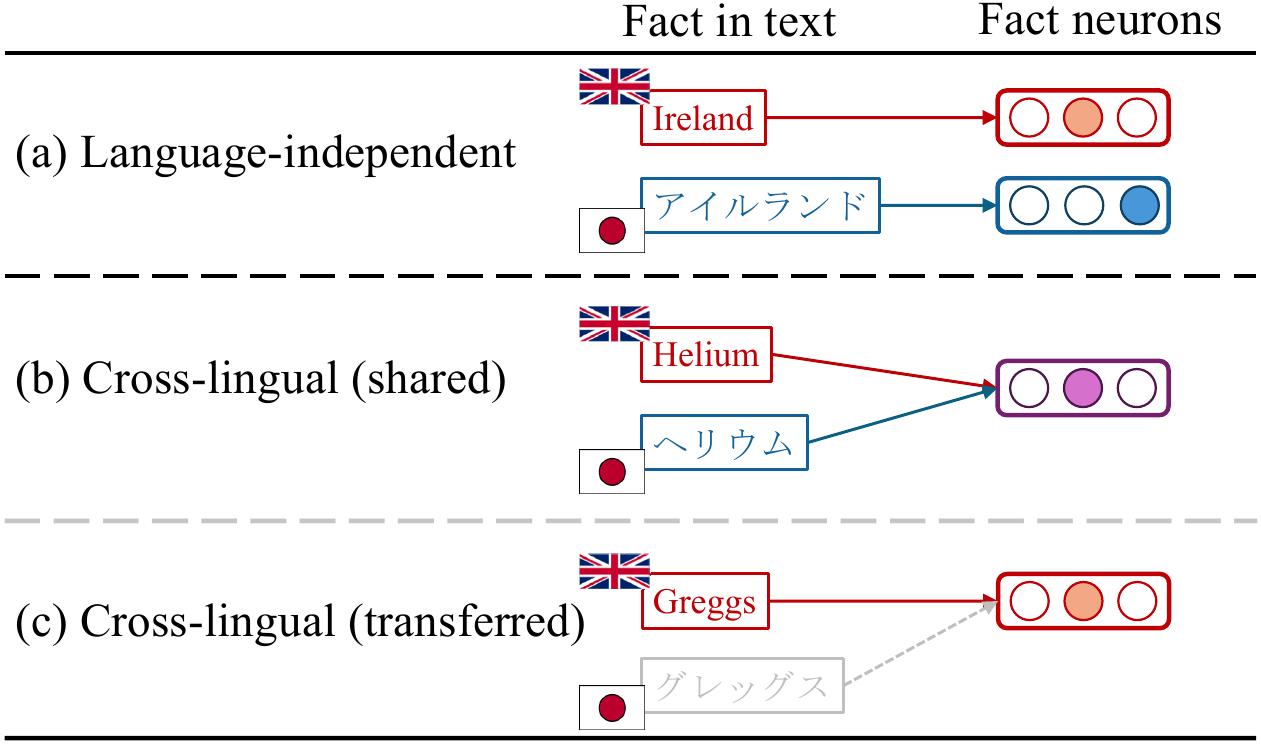}
    \caption{Three types of fact representation in ML-LMs;
    Facts are \yn{a)} represented with distinct neurons across languages (language-independent), \yn{b)} shared using the same neurons (cross-lingual (shared)), and \yn{c)} transferred across languages (cross-lingual (transferred)).}
    \label{fig:factrep}
\end{figure}

\zx{Following early studies~\cite{petroni-etal-2019-language, jiang-etal-2020-know} that \nc{used} cloze-style queries to probe whether monolingual language models can recall 
factual knowledge,
researchers probed ML-LMs~\cite{jiang-etal-2020-x,kassner-etal-2021-multilingual,yin-etal-2022-geomLAMA,fierro-sogaard-2022-factual,keleg-magdy-2023-dlama}.} 
The results indicated that ML-LMs exhibit an ability to recall facts.
However, the mechanism behind the acquisition and representation of facts in ML-LMs remains unclear.

In this study, we investigate\nc{d} whether and how low-resource languages can benefit from the cross-lingual transfer of factual knowledge (Figure~\ref{fig:factrep}). 
Concretely, we 
address\nc{ed} three 
research questions: 

\begin{description}
    \item [RQ1:] How \nc{does} factual probing performance \nc{of ML-LMs} differ across languages, and what factors affect these differences? (§\ref{mtlprobing})
    \item [RQ2:] Do ML-LMs represent 
    \yn{the same fact in different languages with a shared or independent representation?}
    (§\ref{sharing})
    \item [RQ3:] What mechanisms during the pre-training of ML-LMs \nc{affect} the formation of cross-lingual fact representations? (§\ref{transferring})
\end{description}


To 
answer \yn{these} research questions,
we start\nc{ed} by probing \nc{two} ML-LMs, mBERT and XLM-R, using 
the mLAMA probing dataset~\cite{kassner-etal-2021-multilingual}. 
The results reconfirm that ML-LMs \nc{have} difficulty recognizing facts in low-resource languages~\cite{kassner-etal-2021-multilingual}, such as 
Irish and Lithuanian
(§\ref{probing_exp}). 
\yn{However, w}e \yn{also} observed only a 
moderate correlation between probing performance and the amount of training data. Although the cultural bias of \nc{the} mLAMA \nc{dataset} may hinder probing performance in non-Latin script languages~\cite{keleg-magdy-2023-dlama}, the exact \nc{effect} of a model's cross-lingual capabilities remains to be established.

To \nc{identify} the role of cross-lingual capability in fact probing, we perform\nc{ed} a neuron-level analysis for facts predicted correctly. 
By comparing active neurons across languages, we observed that identical facts in various languages are not 
\nc{acquired in identical ways.}
For specific facts, some languages exhibit similar neuron activity, while others display distinct patterns. 
We categorize the former as \textbf{cross-lingual fact representations}, \cshin{as illustrated in Figure~\ref{fig:factrep}(b,c)}, and the latter as \textbf{language-independent representations}, \cshin{as \nc{illustrated} in Figure~\ref{fig:factrep}(a).}

To further discern
cross-lingual representations, we \nc{devised} a 
method \cshin{for tracing the roots of facts} by verifying their presence 
in the knowledge source (Wikipedia for mBERT). 
We assume that the facts that are predicted correctly \nc{although} absent in the training \nc{data} 
are captured through cross-lingual transfer,
referred to as \textbf{cross-lingual transferred}
(Figure~\ref{fig:factrep}(c)) to differentiate from \textbf{cross-lingual shared}
(Figure~\ref{fig:factrep}(b)).
\yn{A} deeper investigation into the results\yn{, however,} revealed that
\cshin{only a fraction of \yn{those facts would be}} acquired \nc{through} cross-lingual transfer. 
This underscores the limitations of current ML-LMs in cross-lingual fact representation.

The contributions of this paper are
\begin{itemize}
\item \yn{Evaluation of} training data volume and mask token count as factors 
to cause discrepancies in probing results across languages and discovery of localized factual knowledge clusters,
\item \cshin{Establishment of
methods for distinguishing among \yn{fact representations in ML-LMs},
by \cshin{identifying shared active neurons and} tracing the roots of facts \yn{back to} the 
training data\nc{, and}}
\item \nc{Revelation} that factual knowledge in ML-LMs 
\nc{has} three types of representations: language-independent, cross-lingual (shared), and cross-lingual (transferred) (Figure~\ref{fig:factrep}). 
\end{itemize}

\section{Related Work}

This section reviews 
existing \yn{studies} on
\cshin{understanding the mechanism of cross-lingual transfer} and factual knowledge probing. 
We first discuss key studies that investigate\nc{d} how knowledge \nc{is transferred} across languages in ML-LMs. Next, we highlight research on how factual knowledge is perceived in pre-trained language models (\cshin{PLMs}).

\subsection{Understanding cross-lingual transfer} 
Numerous \nc{studies} have investigated the \nc{basic} mechanisms of cross-lingual transfer in ML-LMs. 
\nc{Studies of the process of cross-lingual transfer have shown that,} while shared tokens facilitate \cshin{cross-lingual} knowledge transfer, their \nc{effect is} circumscribed~\cite{K2020Cross-Lingual, conneau-etal-2020-emerging}. 
\nc{Subsequent studies showed that using parallel data enhances a model's cross-lingual ability~\cite{moosa-etal-2023-transliteration, reid-artetxe-2023-role}. 
}

\nc{Concurrent studies focused on} the realization of cross-lingual transfer in the parameter space within ML-LMs~\cite{muller-etal-2021-first, chang-etal-2022-geometry, foroutan-etal-2022-discovering}. 
\nc{They reported}
that ML-LMs have both
language-specific and language-agnostic parameter spaces when representing identical \cshin{linguistic} knowledge across 
languages. 
\cshin{However, they focused solely on basic linguistic tasks like dependency parsing and named-entity recognition. Cross-lingual representation of factual knowledge remains underexplored.}
Moreover, while the\nc{se} previous studies primarily \cshin{provide\nc{d} a systematic \nc{explanation} of cross-lingual transfer mechanisms, they \nc{neglected} the detailed variations in how ML-LMs \nc{acquire and represent} specific knowledge.}

\subsection{Factual knowledge probing} 
Understanding factual representation in \cshin{PLMs} has
\nc{attracted much attention recently.} Using fill-in-the-blank cloze question datasets, 
\yn{researchers} explored the \nc{ability} of \cshin{PLMs} \nc{to handle} factual knowledge \nc{in} the English language~\cite{petroni-etal-2019-language, heinzerling-inui-2021-language, wang-etal-2022-finding-skill}.
\nc{To clarify} the mechanism by which \yn{Transformer~\cite{NIPS2017_3f5ee243}}-based \cshin{PLMs} represent facts, \yn{a few} studies have conducted neuron-level investigation~\cite{oba-etal-2021-exploratory,geva-etal-2021-transformer,dai-etal-2022-knowledge}. 
These studies reveal\nc{ed} that specific \nc{fact representation} are linked to a \nc{specific} set of neurons rather than the whole parameter space. This has led to subsequent research focused on enhancing models through neuron adjustments~\cite{de-cao-etal-2021-editing, mitchell2022fast, zhang-etal-2022-moefication}.

\nc{Several studies have investigated the ability of PLMs to represent facts in languages other than English~\cite{jiang-etal-2020-x, kassner-etal-2021-multilingual, fierro-sogaard-2022-factual} in a multilingual setting.}
\nc{Their results} suggest that the ability to perceive factual knowledge is not exclusive to English. 
Other languages demonstrate\nc{d} comparable proficiency. 
However, \nc{weaker} predictability of factual knowledge has been observed for languages with limited resources. 
\nc{One study}~\cite{fierro-sogaard-2022-factual} investigated the \nc{differences in predictability} between languages \nc{and} attributed them to cultural biases.
However, the role of cross-lingual transfer in factual representation across languages has not been extensively explored.

\section{Multilingual Factual Probing} \label{probing_exp}

we \nc{carried} out experiments to probe the factual knowledge of ML-LMs across multiple languages. 
Our objective \nc{was} to clarify how facts are perceived in different languages and to \nc{identify} the difference in fact recognition among languages. 
\yn{Furthermore}, we \nc{investigated} how ML-LMs learn and represent these facts, seeking to understand the interplay between languages in the context of fact recognition.

\subsection{Experiment setup}
\paragraph{Datasets} For the factual probing experiments, we \nc{used} the mLAMA dataset\footnote{While DLama-v1~\cite{keleg-magdy-2023-dlama}, a variant of mLAMA designed to address cultural biases, is available, we opted for mLAMA\nc{, because our focus was} on cross-lingual features rather than solely assessing model competencies in factual understanding. mLAMA is \nc{suitable} for this objective as it offers a consistent query set across all languages, ensuring clarity and precision in our investigation.}~\cite{kassner-etal-2021-multilingual}.
This dataset is a multilingual extension of LAMA~\cite{petroni-etal-2019-language} and draws from sources \nc{such as} TREx~\cite{elsahar-etal-2018-rex} and GoogleRE,\footnote{\url{https://github.com/google-research-datasets/relation-extraction-corpus}} both of which extract information from Wikipedia.
The mLAMA dataset contains 37,498 instances spanning 43 relations, represented as a fill-in-the-blank cloze, \textit{e.g.}, ``[X] plays [Y] music.'' where subject entity X, \yn{a} relation\yn{,} and object \yn{entity} Y form a triplet \cshin{(subject, relation, object)}. 

\paragraph{Models} 
We here focus on probing multilingual factual knowledge using 
encoder-based ML-LMs, 
multilingual BERT (mBERT)\footnote{\url{https://huggingface.co/bert-base-multilingual-cased}}~\cite{devlin-etal-2019-bert} and XLM-R\footnote{\url{https://huggingface.co/FacebookAI/xlm-roberta-large}}~\cite{conneau-etal-2020-unsupervised}.
Encoder-based models are chosen over \cshin{generative models} like mT5~\cite{xue-etal-2021-mt5} \cshin{\yn{and} BLOOM~\cite{workshop2023bloom}}
since they are smaller than \yn{the} generative models but exhibit excellent performance on language understanding tasks.
Specifically for our factual knowledge probing task, which employs fill-in-the-blank queries, \cshin{the encoder-based} models perform well at referencing and integrating information across entire sentences, ensuring a detailed contextual understanding.

\subsection{Evaluation} \label{sec:evaluation}
\cshin{To determine if ML-LMs can capture specific knowledge, we substitute X with the subject tokens and replace Y with mask \yn{tokens} in each cloze template to form \nc{a} cloze query (\textit{e.g.}, ``The Beatles play [MASK] music.''). 
Then, we \nc{instructed} the ML-LMs to predict the mask \yn{tokens}. If, in this instance, it predicted the mask token \nc{to be} ``rock,'' we consider\nc{ed} that the knowledge \nc{was} captured by the ML-LMs.}

\nc{Since the object cannot necessarily be tokenized as a single token, we need to determine the number of mask tokens needed for each probed fact.}
\nc{Previously proposed methods used an} automated technique for determining \nc{the} mask counts \nc{that maximized} the probability of a correct number of mask tokens~\cite{jiang-etal-2020-x, kassner-etal-2021-multilingual}. 
\yn{In contrast, our study aimed at investigating fact representations} rather than \nc{simply} evaluating the probing performance of ML-LMs. \yn{We therefore adopt more lenient methods of probing facts as follows.}

\paragraph{Protocol} To correctly estimate predictable facts,
we evaluate\nc{d} two matching methods: full-match and partial-match. In the full-match approach, we assign\nc{ed} the exact number of mask tokens corresponding to the object. 
However, this method sometimes 
\nc{produced} correct answers containing non-essential tokens such as whitespaces. We 
consider\nc{ed} these cases not as errors but as potentially valid answers.
Consequently, we \yn{also} \cshin{examined} the partial-match method. For each query \yn{template} such as ``[X] plays [Y] music,'' we list\nc{ed} all objects \cshin{Y} and their token counts associated with \yn{the} \yn{template}. We then probe\nc{d} the two ML-LMs \nc{(mBERT and XLM-R)} with multiple queries, ranging from one (\textit{e.g.}, ``\yn{The Beatles} plays [MASK] music'') up to the longest mask token sequence for the corresponding \yn{template} (\textit{e.g.}, ``\yn{The Beatles} plays [MASK] [MASK] [MASK] [MASK] music''). A fact \nc{was} considered correctly predicted if any version of the prompt included the \nc{correct} object tokens, regardless of additional preceding or succeeding tokens. 


\begin{figure}[t]
    \includegraphics[width=\linewidth,clip]{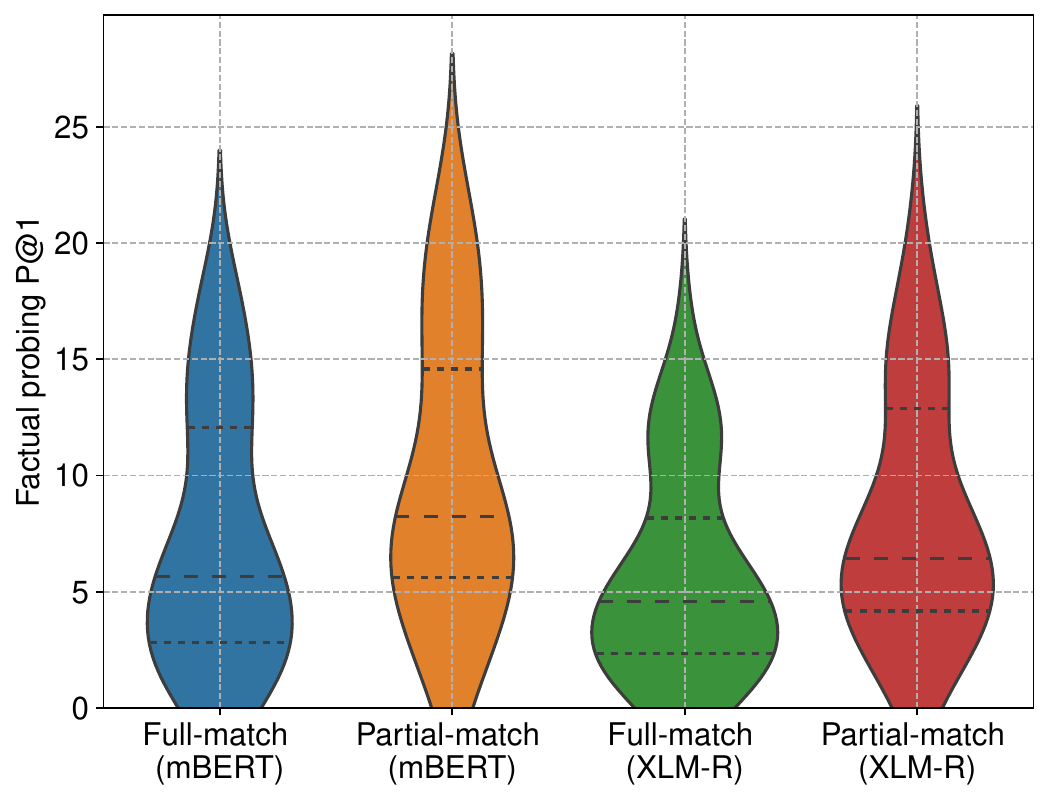}
    \caption{Probing P@1 \cshin{on mLAMA} for full- and partial-match methods with mBERT and XLM-R\@.}
    \label{fig:p1score_vs}
\end{figure}

\paragraph{Results} Figure~\ref{fig:p1score_vs} \nc{displays} the results 
in terms of 
the first rank precision (P@1).
Across all experiments, we noted a consistently low P@1 score, especially for the majority of low-resource languages. 
Refer to Table~\ref{tab:overallp1} in Appendix~\ref{sec:allp1score} for details.

Interestingly, the partial-match method demonstrated noticeably better factual probing by considering partially matched predictions. 
A deeper analysis revealed four unique prediction patterns, specifically discernible \nc{for the} partial-match method, as \nc{examplified} in Table~\ref{tab:extra_tokens}. 
These patterns \nc{illustrate} the \nc{limitation} of the factual probing based on the fill-in-the-blank dataset\nc{: the \yn{answers} are restricted to a single standard format and thus do not reflect the diversity in entity expressions in text.}
These observations indicate a direction for future improvements in probing techniques.

For clarity in our subsequent analysis, we will primarily focus on mBERT, a 12-layer Transformer\yn{-based} \yn{ML-LM pre-}trained on Wikipedia text across 103 languages. This decision is motivated by the comparable results between mBERT and XLM-R\yn{\@}. Although the partial-match method offers a richer representation for exploration, it sometimes includes irrelevant tokens that can introduce noise \cshin{(\textit{e.g.}, whitespace in Table~\ref{tab:extra_tokens})}. Therefore, the following discussions \nc{are} predominantly based on results obtained using the full-match approach.

\begin{table}[t]
    \centering
    \small
    \tabcolsep1.1pt 
    \begin{tabular}{ll}
    \toprule
    \textbf{Type} & \textbf{Example}\\
    \midrule
    Whitespace & Petr Kroutil was born in Prague ( ). \\
    Preposition & Galactic halo is part of (the) galaxy. \\
    Related noun & Surinder Khanna was born in Delhi (,) (India). \\
    Adjective & Pokhara Airport is a (popular) airport.  \\
    \bottomrule
    \end{tabular}
    \caption{\nc{Four patterns discerned in facts predicted by partial-match method.} The tokens in ``()'' are extra compared \nc{with those in the ground-truth dataset}.} 
    \label{tab:extra_tokens}
\end{table}

\section{What Factors Influence Discrepancy in Factual Probing across Languages?} \label{mtlprobing}

\begin{figure*}[t]
    \centering
    \includegraphics[width=\textwidth,clip]{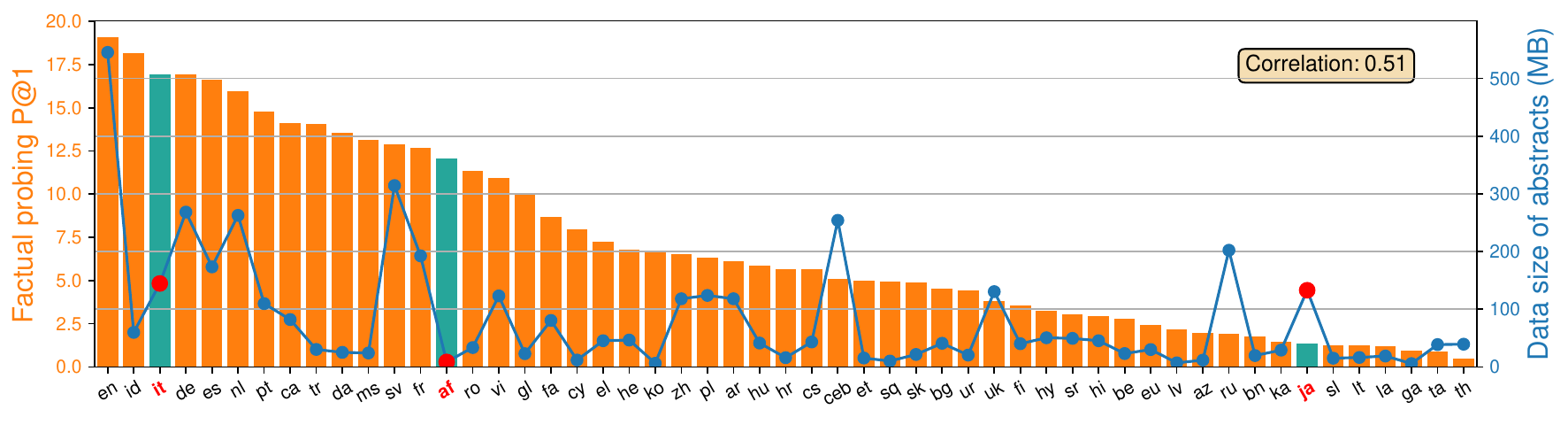}
    \caption{Wikipedia data size of abstracts vs.\ Factual probing P@1 \yn{on mLAMA in mBERT in} 53 languages.}
    \label{fig:pagevsp1}
\end{figure*}

\cshin{The vertical bars in Figure}~\ref{fig:pagevsp1} \nc{showing} the results of 
factual probing in various languages\nc{reveal substantial differences} among languages. 
In this section, we will \yn{evaluate} the potential factors for \nc{these differences} and 
\cshin{examine how \nc{they} relate to proficiency in the cross-lingual transfer ability}
of ML-LMs.

\subsection{\yn{Training data volume for learning facts}}\label{subsec:data_accu}
The first factor \yn{relates to} the amount of distinct factual knowledge seen in training the ML-LMs. Since it is difficult to estimate the amount of factual knowledge in the training data, we explored several metrics on 
\cshin{the training data volume instead. Specifically, we calculated the Pearson correlation coefficient between probing accuracy (P@1) and five metrics on the training data of  
mBERT, Wikipedia:\footnote{\label{fn:wikidumps}We crawled Wikipedia dumps as of the time just before the released date of mBERT\@. See Appendix~\ref{sec:wikidumps} for details.}}
the number of Wikipedia articles
and raw and compressed data sizes for abstracts and full articles.

\begin{table}[t]
\small
\tabcolsep 2pt
\begin{tabular}{lr}
\toprule
\textbf{Statistics} & \textbf{Pearson's $zr$ with P@1} \\
\midrule
The number of page count & 0.43 \\
The data size of articles & 0.44 \\
The data size of articles (bzipped) & 0.45 \\
The data size of abstracts & \textbf{0.51} \\
The data size of abstracts (bzipped) & 0.48 \\
\bottomrule
\end{tabular}
\caption{Correlation \yn{between the training data volume and probing P@1 on mLAMA with mBERT\@.}}\label{fig:corr_size_p1}
\end{table}

\cshin{Table~\ref{fig:corr_size_p1} lists the correlation between P@1 and the metrics on the training data volume. All metrics show a moderate correlation with P@1\@.
We depict the data size of abstracts, which correlated the most among the five metrics,
in Figure~\ref{fig:pagevsp1}.
The moderate \yn{correlation} indicates a limited impact of the training data volume on learning factual knowledge.}

\begin{table}[t]
    \small
    \centering
    \begin{tabular}{lccc}
    \toprule
    \textbf{} & \textbf{it} & \textbf{ja} & \textbf{af}\\
    \midrule
    mBERT P@1 & 16.94 & 1.34 & 12.05\\
    One-token P@1& 15.27 & 15.34 & 17.00 \\
    One-token entities & 1675 & 126 & 498 \\
    \midrule
    XLM-R P@1 & 10.80 & 4.78 & 8.17\\
    One-token P@1 & 13.67 & 14.73 & 16.58 \\
    One-token entities & 923 & 244 & 333 \\
    \bottomrule
    \end{tabular}
    \caption{P@1 and one-token object counts for mBERT and XLM-R in Italian (it), Japanese (ja) Afrikaans (af).}
    \label{tab:subwordnumber}
\end{table}
\cshin{Among high-resource languages,
we observed the \yn{probing P@1} of 16.94\% for Italian (it) and 1.34\% for Japanese (ja), as shown in Table~\ref{tab:subwordnumber}.}
Prior research has highlighted potential cultural biases in mLAMA, particularly \nc{affecting} non-Latin script languages~\cite{keleg-magdy-2023-dlama}. 
However, these biases alone do not explain the \nc{substantial difference} between the 
training data volume
and probing performance. 
Meanwhile, some low-resource languages, such as Afrikaans (af), perform 
relatively well despite having limited Wikipedia data.
The ability of Afrikaans (af) to represent such a breadth of knowledge, even in the face of potential cultural biases, is indeed remarkable. 


\subsection{\yn{Mask token count making inference hard}}\label{subsec:mask}
There \nc{was} \yn{notable $-0.81$ (mBERT) and $-0.74$ (XLM-R) correlations between P@1 and the number of subwords in the target entities.} 
\yn{As shown in Table~\ref{tab:subwordnumber}}, while both ML-LMs had similar P@1 scores in predicting one-token entities, 
the XLM-R tokenizer captured more one-token entities in Japanese (ja), resulting in more accurate predictions. 
The XLM-R tokenizer often produced shorter tokens for non-Latin scripts, enhancing its performance for non-Latin languages.
However, this does not 
explain the differences in prediction accuracy across languages,
as Afrikaans (af) 
outperformed Japanese (ja) for one-token P@1.

\begin{figure}[t]
    \centering
    \includegraphics[width=\linewidth,clip]{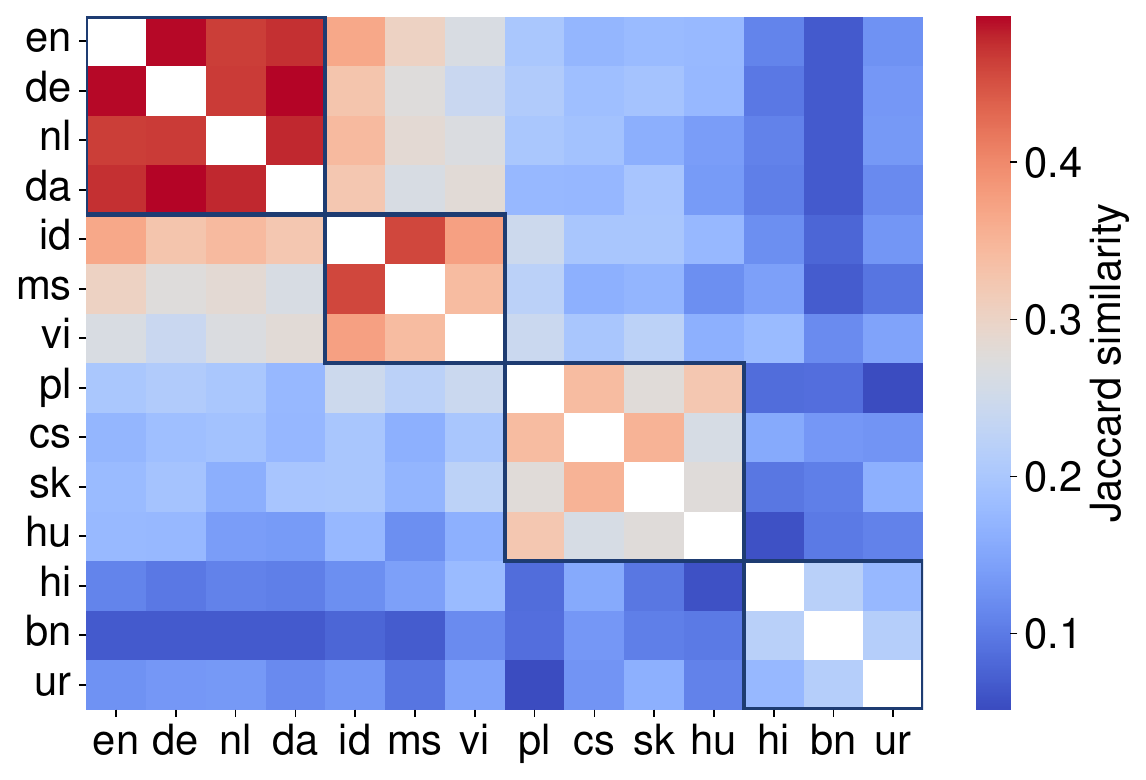}
    \caption{Jaccard similarity matrix of shared factual knowledge across languages \nc{with} \yn{mBERT}.}
    \label{fig:langzone}
\end{figure}

\subsection{Presence of localized knowledge cluster}
\label{subsec:localized_knowledge}
The higher accuracy \nc{for} low-resource languages might \nc{have resulted} from the \nc{model being proficient at} cross-lingual factual knowledge sharing. 
To \nc{investigate} this \nc{possibility}, we assessed shared facts between languages using Jaccard similarity:
\begin{align} 
J(A, B) &= \frac{|A \cap B|}{|A \cup B|},
\end{align}
where \(A\) and \(B\) are sets of facts predictable by two languages. 

Figure~\ref{fig:langzone} reveals that languages in geographical proximity show\nc{ed} greater overlap in shared facts.
Geographically \nc{proximate} languages, like Indonesian (id), Malay (ms), and Vietnamese (vi),
had higher similarities, \nc{indicating} substantial shared content.
This suggests that 
cross-lingual knowledge transfer does not occur universally across languages.
Instead, it appears to be localized, driven more by shared culture and vocabulary. We will explore this phenomenon in the subsequent sections. 

\paragraph{Summary} We examined \yn{training data volume and mask token count as factors} that 
will
influence the discrepancies in factual knowledge comprehension across languages.
Our findings reveal\nc{ed} localized knowledge sharing patterns among languages, hinting at the potential for cross-lingual transfer. 

\section{\yn{Do ML-LMs Have Fact Representations Shared across Languages?}} \label{sharing}

In this section, we \nc{discuss} \cshin{how ML-LMs represent facts within their parameter spaces by exploring two scenarios.}
\nc{In one scenario, a copy of the same fact is \yn{independently} maintained in different languages, as illustrated in Figure~\ref{fig:factrep}(a); ML-LMs based on this scenario are referred to as ``language-independent.''}
\nc{In the other scenario, fact representations in different languages are unified in an embedding space, as illustrated in Figure~\ref{fig:factrep}(b,c); ML-LMs based on this scenario are referred to as ``cross-lingual.''}
\yn{The language-independent scenario will hinder cross-lingual transfer in fine-tuning ML-LMs on downstream tasks, where the training data is available in a few languages.}

\subsection{Factual neuron probing} \label{4.1}

In Transformer-based PLMs, the feed-forward network (FFN) plays a pivotal role in the knowledge extraction and representation process~\cite{durrani-etal-2020-analyzing, dai-etal-2022-knowledge}. 
Formally, an FFN is defined as: 
\begin{equation}
\text{FFN}(x) = \text{f}(x\mathbb{K}^T + b_1) \mathbb{V} + b_2
\end{equation}
where \(\mathbb{K}\), \(\mathbb{V}\), \(b_1\), and \(b_2\) are trainable parameters. 


\paragraph{Experiment setup} We analyzed the representation of cross-lingual facts in ML-LMs by identifying their active neurons across languages. We \nc{used} \cshin{a method called PROBELESS}~\cite{antverg2022on} - an efficient and explicit technique that measures neuron activity by contrasting value differences among facts. Specifically, \cshin{PROBELESS} identifies neurons as active when their values deviate greatly from the average for specific knowledge representations.

\nc{More specifically, we analyzed} neuron activity for each correctly-predicted fact, represented as (subject, relation, object). For probing, we consider\nc{ed} other predictable facts that share the same relation but vary in subject-object pairs. We collect\nc{ed} the neurons of the mask tokens and identified their active neurons as \yn{signatures} of the fact \yn{representation}s. For multi-\yn{token} masks, we use\nc{d} average pooling across all tokens. 
As our goal \nc{was} to investigate fact representations across languages, we collect\nc{ed} the active neurons for the same fact in various languages for further analysis. \nc{Because} the reliability of fact probing \nc{is lower} when \nc{the availability of predicted facts is limited}, 
we focused on the top 30 languages by P@1 score.

\begin{figure}[t]
    \centering
    \includegraphics[width=\linewidth,clip]{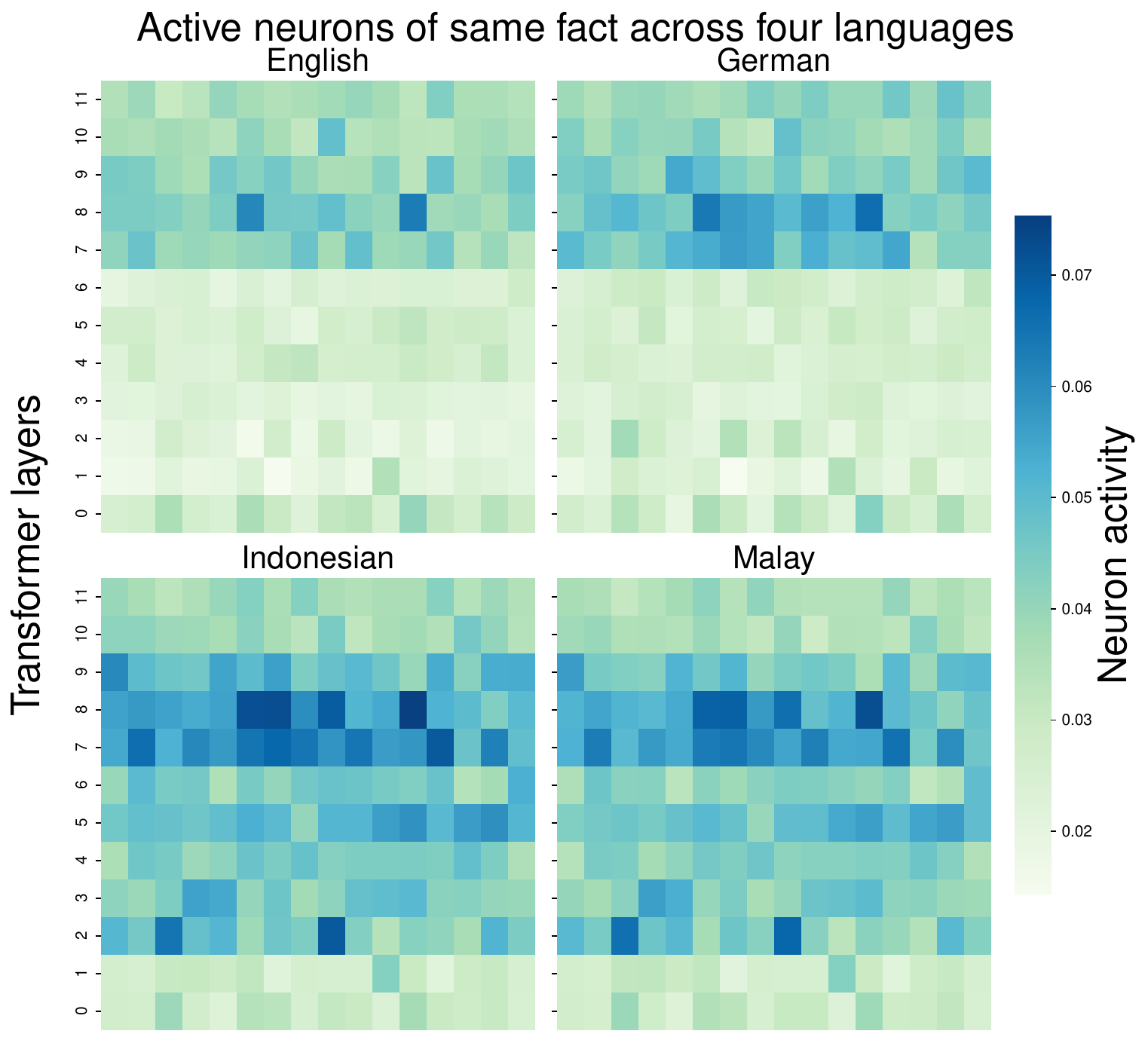}
    \caption{Neuron activity with mBERT in four languages, English, German, Indonesian, and Malay, in response to the query ``William Pitt the Younger used to work in [MASK].'' Color intensity indicates neuron activity; neurons in each Transformer layer are grouped into 16 bins. Distinct activation patterns in the English-German and Indonesian-Malay pairs indicate cross-lingual knowledge neurons, while differences between the pairs indicate language-independent representations.}
    \label{fig:knowledge_neurons}
\end{figure}

\begin{figure}[t]
    \includegraphics[width=\linewidth,clip]{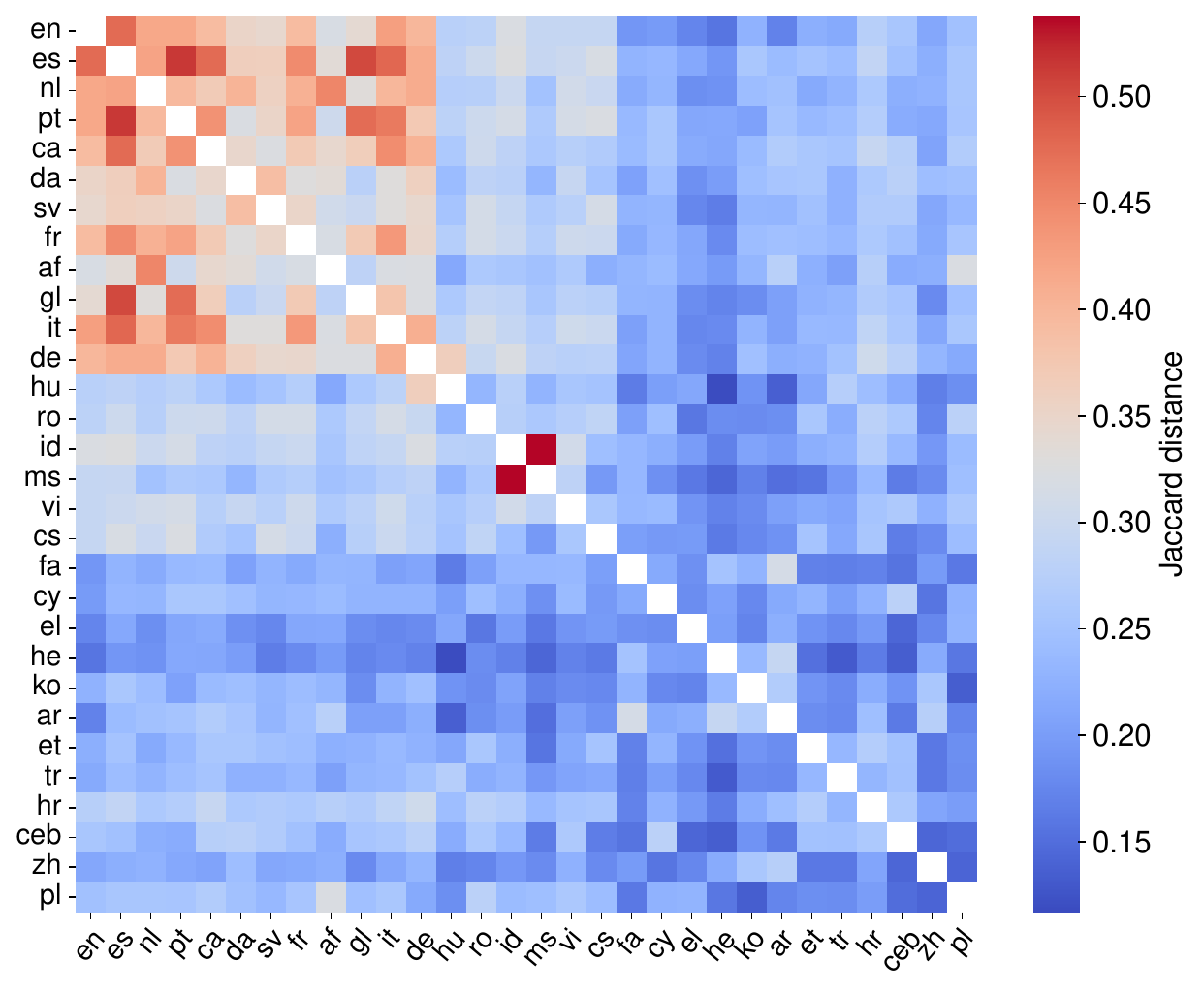}
    \caption{\yn{Language} similarity \nc{based on} top 50 shared active neurons by probing on mLAMA \yn{with mBERT}\@.}
    \label{fig:lang_pairwise_dist_by_shared_neurons}
\end{figure}

\subsection{Results and discussion} \label{result_discussion}
\paragraph{\yn{Do 
cross-lingual fact representations exist?}} In our neuron probing, we \nc{identified and used} active neurons to \nc{distinguish} \yn{cross-lingual fact representations from language-independent neurons.} Similar patterns in active neurons across languages suggest that there is \nc{a common} cross-lingual semantic space for fact representation.
Our findings indicate that while some languages exhibit similar neuron activity patterns for a given fact, others exhibit distinct distributions, as depicted in Figure~\ref{fig:knowledge_neurons}. 
This \nc{indicates} the presence of both language-independent and cross-lingual fact representations in ML-LMs, even for the same fact.

\begin{figure*}[t]
    \centering
    \includegraphics[width=\textwidth,clip]{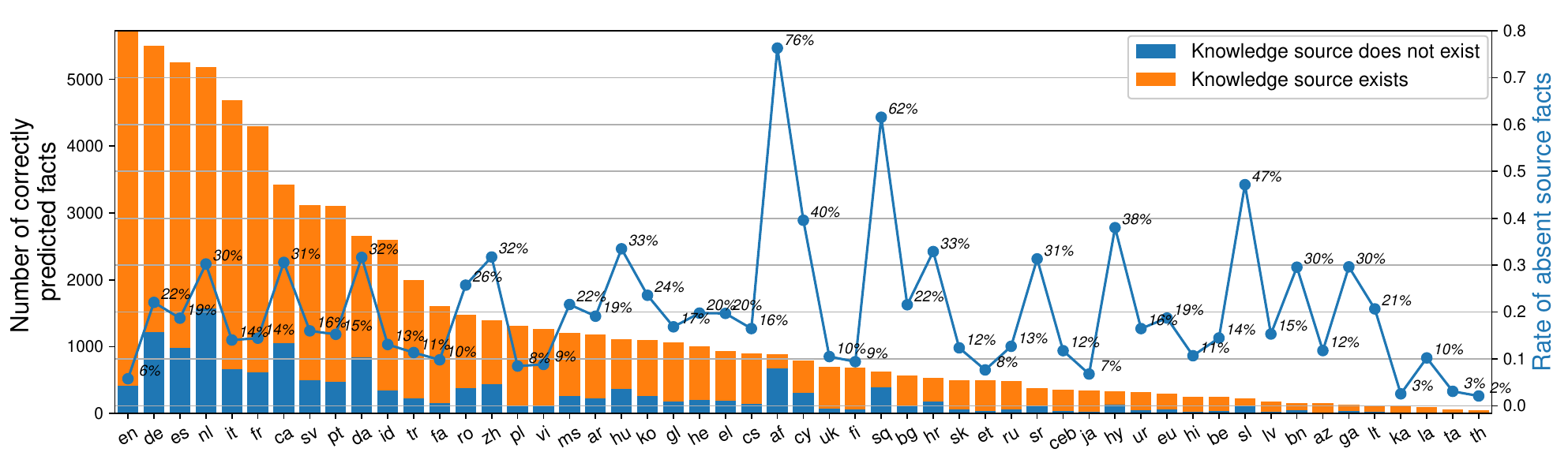}
    \caption{Number of correctly-predicted facts \nc{with mBERT} in terms of existence of knowledge source.}
    \label{fig:absentrate}
\end{figure*}

\paragraph{\nc{Quantification of} cross-lingual sharing} 
To precisely measure the extent of cross-lingual sharing of facts between two languages, we \nc{calculated} Jaccard similarity \nc{based on} the top 50 active factual neurons. 
We then measure\nc{d} the general language similarity \nc{among} all languages by computing the average similarity for all shared facts.

\yn{Figure~\ref{fig:lang_pairwise_dist_by_shared_neurons} shows the results \yn{of computing general language similarity in terms of shared facts}.} Surprisingly, our findings revealed no consistent geographical boundaries among languages, suggesting that \nc{the use of} either the language-independent \nc{scenario} \nc{or the} cross-lingual sharing \nc{scenario}  \cshin{largely depends on the specific factual knowledge itself}, \nc{so} such analysis should be tailored to specific factual knowledge. 
For instance, despite English (en) and Chinese (zh) exhibiting a relatively low neuron correlation (\yn{$0.21$}, compared \nc{with} the \yn{$0.24$} average), they still display\nc{ed} similar active neuron \nc{patterns} for certain facts, often rooted in shared tokens, like ``Google'' in Chinese ``developed-by'' relations. 


The results of neuron probing 
revealed that active fact neurons in low-resource languages have more activity and are distributed more in the shallow layers of Transformers compared with high-resource languages. This finding contradicts \nc{the findings of} previous research~\cite{oba-etal-2021-exploratory, dai-etal-2022-knowledge}, \nc{and} suggests that only a few neurons in \nc{the} higher Transformer layers are responsible for representing facts. This \nc{difference} indicates a potential reason for the lower expression ability of low-resource languages, \nc{for which} the hierarchical structure of knowledge is not acquired as well as in resource-rich languages.

\paragraph{Summary}
\nc{Our exploratory analysis using neuron probing of fact representation in ML-LMs and examination of whether languages share common representations or maintain unique knowledge spaces for specific facts revealed \yn{the presence of} both language-independent and cross-lingual neural activity patterns across languages. The results of Jaccard similarity analysis of active factual neurons revealed inconsistent geographical boundaries in knowledge sharing, indicating the complexity of cross-lingual knowledge representation.}

\section{How Are Cross-lingual Representations of \yn{Facts} Formed \yn{in ML-LMs}?}\label{transferring}

\nc{Having confirmed} the presence of cross-lingual representations, we subsequently explore\nc{d} \nc{their} formation within ML-LMs \nc{and assessed} whether they are learned individually from distinct language corpora and subsequently aligned into a common semantic space
(Figure~\ref{fig:factrep}(b))
or \nc{whether} they are acquired through cross-lingual transfer
(Figure~\ref{fig:factrep}(c)). 

\subsection{Tracing the roots of facts \yn{back to data}} \label{subsec:tracingfacts}

To identify the reason behind the formation of \nc{a} cross-lingual representation, it is crucial to verify if the fact originates from the training data. We \nc{used} a simple yet effective method to check the presence of a fact in text: for a fact triplet (subject, relation, object), we examine\nc{d} the occurrence\nc{s} of the subject and object in \yn{mBERT's} training data, Wikipedia.\footnoteref{fn:wikidumps} If they \nc{could be} found, the fact \nc{was} considered present \nc{in the \yn{data}}. Although this approach may not provide precise quantitative results, it helps in exploring cross-lingual transfer possibilities. 



To 
determine \nc{whether}
a fact \nc{was} 
traced back to \yn{the data}, we \nc{used} subject-object co-occurrence as an approximation method. We rigorously adhere\nc{d} to the preprocessing and sentence-splitting guidelines for mBERT~\cite{devlin-etal-2019-bert}. 
Using the WikiExtractor,\footnote{\url{https://github.com/attardi/wikiextractor}} we extracted only text passages, deliberately omitting lists, tables, and headers. Each extracted document \nc{was} segmented into multiple lines, with each line containing no more than 512\footnote{\yn{The maximum number of tokens \nc{that can be} input to mBERT in training.}} tokens. \nc{Using} string matching between the object\yn{-}subject \nc{pair} and Wikipedia \nc{text}, we assess\nc{d} the co-occurrence of the object and subject for a given fact. If \nc{there was co-occurrence}, we consider\nc{ed} the fact to be present; \nc{otherwise, it was considered to be absent.}

\subsection{Analysis of absent \yn{yet predictable} facts} \label{analysis_absent_facts}

We assessed 
\yn{the absence rate of all and correctly predicted facts, respectively.} 
\nc{As shown by the results for 53 languages in Figure~\ref{fig:absentrate}, languages with more training data exhibited better factual knowledge coverage, as anticipated.}
Nonetheless, several facts, such as those in Afrikaans (af) \yn{and Albanian (sq)}, \nc{were} accurately predicted even without verifiable existence in the training corpus, \nc{indicating} a high possibility of \nc{effective} cross-lingual transfer.

\paragraph{\yn{What kinds of facts are absent yet predictable?}}
\nc{Analysis revealed} that many of the facts that were absent in the knowledge source but correctly predicted \nc{were} relatively easy to predict. We categorized \yn{these easy-to-predict facts} into two types: \cshin{shared entity tokens and naming cues}. \nc{Along with other} facts, we grouped them into a total of three categories \nc{by using} a rule-based method \cshin{(see to Appendix~\ref{sec:factcheck} for the criteria of fact classification).}

\begin{description}

\item[Shared entity tokens:] Some probing queries ask object entities whose tokens are 
shared with the subject entities; for example,
`Sega Sports R\&D is owned by Sega.' We regard correctly predicted facts \nc{to be of} this type when 
the subject and object entities share subwords.

\item[Naming cues:] 
Some probing queries \nc{are related} to entity-universal association across person names, countries, and languages  (see Table~\ref{tab:naming_rels} in Appendix for details), which \yn{allows} the ML-LMs to guess the object entity from subwords of the subject entity; for example, `The native language of Go Hyeon-jeong is Korean.' 
\nc{We regard facts correctly-predicted on the basis of such a relation to be of this type}.

\item[Others:] The remaining facts are 
difficult to infer from the entities only, \nc{indicating} the high possibility of cross-lingual transfer. \textit{e.g.}, `Crime \& Punishment originally aired on NBC\@.'

\end{description}

\begin{figure}[t]
    \centering
    \includegraphics[width=\linewidth,clip]{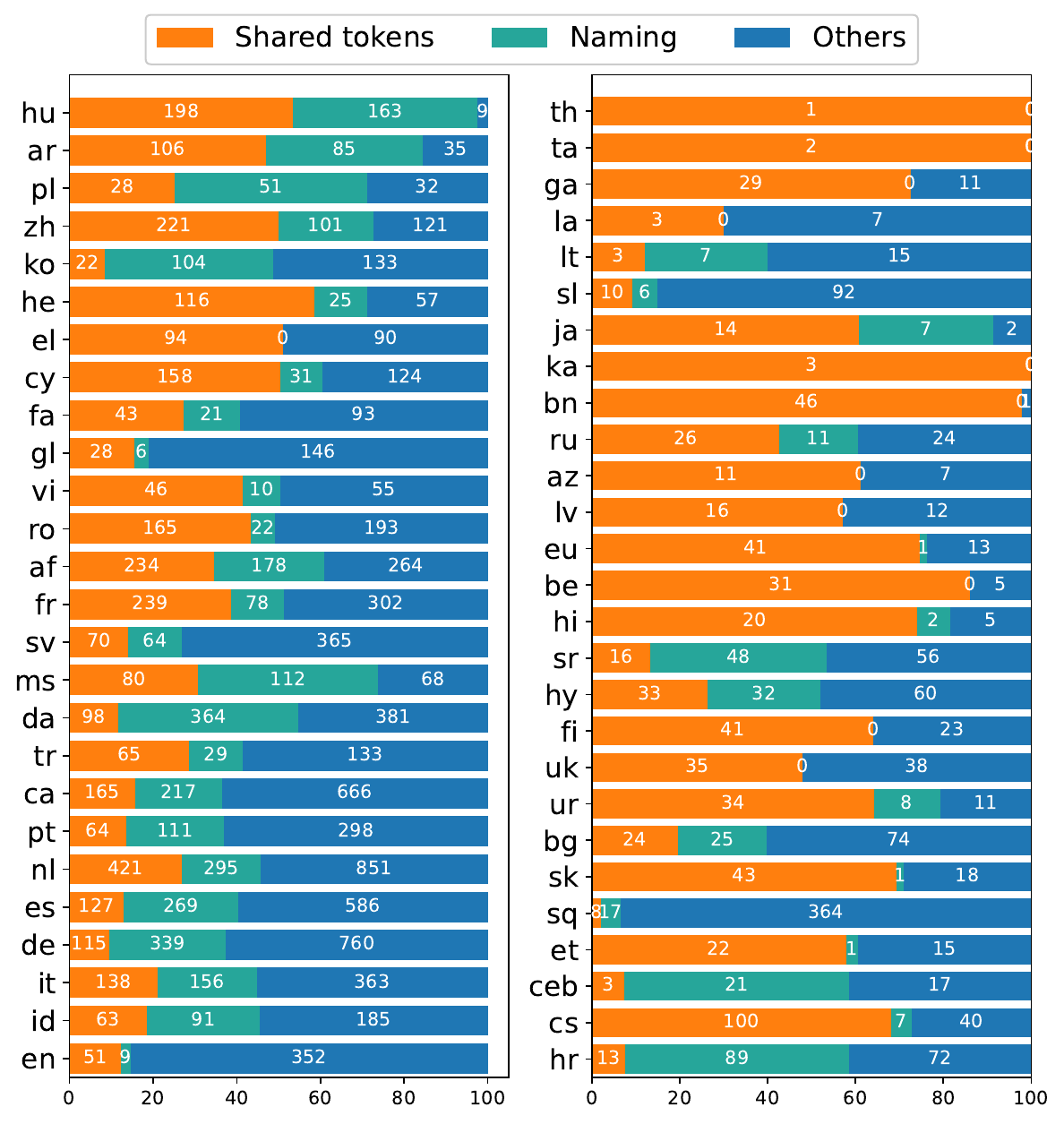}
    \caption{The count of three types of absent and predictable facts with mBERT\@.}
    \label{fig:3types}
\end{figure}

Figure~\ref{fig:3types} shows the \nc{proportions} of \nc{facts correctly predicted without \yn{knowledge sources}} by mBERT \nc{for the three types}.
The predictability of easy-to-predict facts suggests that \cshin{the ML-LMs} can rely on \nc{simple} deductions rather than encoding specific facts to make predictions, highlighting the need to enhance 
probing datasets to \nc{enable} more effective evaluation of model proficiency in fact representation. 
\nc{Without} the easy-to-predict facts, the absence rate drops but \nc{is} still not zero (blue bar in Figure~\ref{fig:3types}) for some of the languages, such as Albanian (sq), Slovenian (sl), and \cshin{Galician (gl)}, indicating that ML-LMs indeed possess cross-lingual transfer capabilities for factual knowledge, \nc{even though the knowledge sources for some languages are limited}.
See Table~\ref{table:aaa} through \ref{table:ccc} in Appendix for 
examples \yn{of} correctly-predicted facts of the three types.

\paragraph{Summary} \cshin{We trace the roots of facts back to the
mBERT's pre-training data, specifically the Wikipedia \yn{text}. We categorized correctly predicted but absent facts into three types, two of which can be resolved through simple inference. Our statistics show that while cross-lingual transfer of factual knowledge in ML-LMs does occur, it is limited, highlighting the challenges in achieving effective cross-lingual factual knowledge transfer.} 

\section{Conclusions}

Our research establishes the groundwork for further studies in understanding cross-lingual factual knowledge representation. 
\cshin{Through comprehensive factual probing experiments and analysis \nc{for} 53 languages using the mBERT \nc{multilingual language model}, we \yn{evaluated} key factors \nc{in the differences between their proficiencies in cross-lingual transfer of factual knowledge}, such as \yn{the training} data volume and mask token count, and identified knowledge sharing patterns among \nc{geographically proximate} language clusters.}

\cshin{We leverage the existing neuron probing and the proposed knowledge tracing methods} to identify three \nc{types of patterns} for acquiring and representing factual knowledge across languages in ML-LMs: language-independent, cross-lingual shared, and cross-lingual transferred. 
Analysis revealed the challenges involved in achieving effective cross-lingual transfer of factual knowledge from high-resource to low-resource languages in ML-LMs. 

\nc{Future work aims to} enhance the cross-lingual fact representation learning in ML-LMs and develop a more precise factual probing dataset.

\section{Limitations}
We primarily examined two encoder-based 
models for language understanding tasks, mBERT and XLM-R\@. 
Therefore, our findings may not directly apply to the recent, \yn{large} decoder-based multilingual language models such as BLOOM~\cite{workshop2023bloom}.
Future research \nc{is needed to} explore these larger generative models \nc{in order} to gain more insights into \nc{the} mechanism of cross-lingual knowledge transfer in ML-LMs.

Moreover, the dataset we \nc{used} has certain limitations. \cshin{\nc{A review of the relation} template in mLAMA by one \cshin{first} author, who is a native Chinese speaker, identified necessary corrections for certain Chinese language prompts}. Meanwhile, the dataset focuses on a limited set of relation types, indicating that fact prediction \cshin{in other relations} may lie beyond the scope of our current research.

\section{Ethical Statement}
All datasets \nc{used} \cshin{in our experiment} are publicly accessible and do not contain sensitive information.\footnote{\url{https://en.wikipedia.org/wiki/Wikipedia:What_Wikipedia_is_not}} The findings and interpretations presented are unbiased and intended for academic purposes.

\section*{Acknowledgements}
This work was partially supported by the special fund of Institute of Industrial Science, The University of Tokyo, by JSPS KAKENHI Grant Number JP21H03494, JP21H03445, and by JST, CREST Grant Number JPMJCR19A, Japan.

\bibliography{acl_latex}

\appendix

\begin{table*}[t]
    \centering
    \small
    \tabcolsep 5pt
    \begin{tabular}{clrrrr@{\qquad\qquad}clrrrr}
    \toprule
        \multirow{2}{*}[-3pt]{\textbf{ISO}} &\multirow{2}{*}[-3pt]{\textbf{Language}} & \multicolumn{2}{c}{\textbf{mBERT}} & \multicolumn{2}{c@{\quad\qquad}}{\textbf{XLM-R}} & \multirow{2}{*}[-3pt]{\textbf{ISO}} & \multirow{2}{*}[-3pt]{\textbf{Language}} & \multicolumn{2}{c}{\textbf{mBERT}} & \multicolumn{2}{c}{\textbf{XLM-R}} \\
        \cmidrule(lr){3-4}
        \cmidrule(lr{3em}){5-6}
        \cmidrule(lr){9-10}
        \cmidrule(lr){11-12}
                & & \textbf{Full} & \textbf{Partial} & \textbf{Full} & \textbf{Partial}  & & & \textbf{Full} & \textbf{Partial} & \textbf{Full} & \textbf{Partial} \\
        \midrule
        en & English & 19.07 & \textbf{22.57} & 17.08 & 21.17 & cs & Czech & 5.63 & \textbf{8.62} & 1.21 & 4.34 \\
        id & Indonesian & 18.15 & \textbf{22.43} & 13.99 & 19.23 & ceb & Cebuano & 5.11 & \textbf{5.84} & 0.76 & 0.88 \\
        it & Italian & 16.94 & \textbf{19.78} & 10.80 & 13.53 & et & Estonian & 4.97 & \textbf{8.24} & 3.82 & 6.01 \\
        de & German & 16.91 & \textbf{20.33} & 12.06 & 14.78 & sq & Albanian & 4.93 & \textbf{5.62} & 3.31 & 4.13\\
        es & Spanish & 16.65 & \textbf{20.28} & 10.51 & 12.87 & sk & Slovak & 4.90 & \textbf{7.08} & 2.84 & 4.84 \\
        nl & Dutch & 15.98 & \textbf{18.30} & 10.47 & 13.04 & bg & Bulgarian & 4.51 & 6.58 & 5.07 & \textbf{7.44} \\
        pt & Portuguese & 14.76 & \textbf{17.96} & 14.05 & 17.12 & ur & Urdu & 4.41 & \textbf{8.02} & 4.40 & 6.31\\
        ca & Catalan & 14.11 & \textbf{17.05} & 5.23 & 8.60 & uk & Ukrainian & 3.84 & \textbf{6.56} & 0.64 & 4.18  \\
        tr & Turkish & 14.08 & \textbf{17.65} & 13.79 & 17.47 & fi & Finnish & 3.58 & 7.11 & 4.43 & \textbf{8.54} \\
        da & Danish & 13.56 & \textbf{16.61} & 12.01 & 15.63  & hy & Armenian & 3.25 & \textbf{5.01} & 3.90 & 4.66 \\
        ms & Malay & 13.14 & \textbf{16.99} & 11.20 & 14.76 & sr & Serbian & 3.07 & \textbf{5.95} & 2.45 & 5.59\\
        sv & Swedish & 12.89 & \textbf{15.32} & 11.63 & 13.63 & hi & Hindi & 2.95 & 5.63 & 3.78 & \textbf{6.61} \\
        fr & French & 12.68 & \textbf{20.18} & 7.79 & 13.81 & be & Belarusian & 2.80 & \textbf{4.49} & 0.78 & 1.54\\
        af & Afrikaans & 12.05 & \textbf{14.47} & 8.17 & 10.09  & eu & Basque & 2.45 & \textbf{5.42} & 1.19 & 2.46\\
        \yn{ro} & Romanian & 11.33 & 14.23 & 13.38 & \textbf{17.46} & lv & Latvian & 2.15 & \textbf{3.79} & 1.66 & 2.94 \\
        vi & Vietnamese & 10.93 & 14.58 & 11.78 & \textbf{15.67}  & az & Azerbaijani & 1.99 & 5.60 & 3.21 & \textbf{6.38} \\
        gl & Galician & 10.00 & \textbf{13.03} & 6.04 & 8.00 & ru & Russian & 1.90 & \textbf{5.98} & 0.79 & 4.07 \\
        fa & Persian & 8.67 & \textbf{12.47} & 7.30 & 9.36 & bn & Bangla & 1.76 & 3.12 & 2.67 & \textbf{4.10} \\
        cy & Welsh & 7.98 & \textbf{9.16} & 5.08 & 6.05 & ka & Georgian & 1.45 & 1.79 & 1.89 & \textbf{2.31}  \\
        el & Greek & 7.24 & \textbf{8.17} & 5.68 & 7.41 & ja & Japanese & 1.34 & 4.85 & 4.78 & \textbf{5.26} \\
        he & Hebrew & 6.78 & \textbf{9.09} & 4.60 & 6.44 & sl & Slovenian & 1.26 & \textbf{3.80} & 1.77 & 3.70 \\
        ko & Korean & 6.73 & \textbf{9.24} & 7.18 & 6.44 & lt & Lithuanian & 1.25 & 1.94 & 2.31 & \textbf{3.42} \\
        zh & Chinese & 6.51 & \textbf{11.95} & 4.05 & 5.91 & la & Latin & 1.21 & 2.24 & 1.83 & \textbf{2.53} \\
        pl & Polish & 6.33 & \textbf{8.45}& 5.09 & 8.30 & ga & Irish & 0.96 & \textbf{1.31} & 0.56 & 0.75 \\
        ar & Arabic & 6.11 & \textbf{8.25} & 6.16 & 7.63 & ta & Tamil & 0.90 & \textbf{1.93} & 0.93 & 1.24  \\
        hu & Hungarian & 5.86 & 10.08 & 5.42 & \textbf{11.17}  & th & Thai & 0.49 & 0.65 & 2.75 & \textbf{4.26} \\
        hr & Croatian & 5.65 & \textbf{9.51} & 2.36 & 5.27 & \multicolumn{2}{l}{Macro average} & 8.85 & \textbf{11.84} & 6.88 & 9.52 \\
         \bottomrule
         \end{tabular}
     \caption{\yn{P@1 \yn{for 53 languages on mLAMA} using full- and partial-match methods with mBERT and XLM-R\@}.}
     \label{tab:overallp1}
 \end{table*}

    \section{\yn{Probing P@1 on mLAMA with mBERT and XLM-R}}\label{sec:allp1score}

Table~\ref{tab:overallp1} lists the probing P@1 \yn{for the 53 languages on mLAMA} using full-match and partial match methods with mBERT and XLM-R, respectively, \cshin{to complement the overall results shown in \S~\ref{sec:evaluation}. In most of the languages, mBERT with the partial-match method achieved the best P@1\@.}
This is probably because the number of facts in Wikipedia used in mBERT will be larger than that in CC-100 used in XLM-R, although CC-100 is larger than Wikipedia~\cite{conneau-etal-2020-unsupervised}. Meanwhile, XLM-R outperforms mBERT on languages with non-Latin scripts such as Hindi, Bangla, Georgian, Japanese, and Thai and several Eastern European languages such as Romanian, Hungarian, Bulgarian, Finnish, Azerbaijani, and Georgian. These results will be probably due to the larger mask token count in the non-Latin scripts (\S~\ref{subsec:mask}) and local knowledge clusters (\S~\ref{subsec:localized_knowledge}) \yn{that do not include resource-rich languages.}

\section{Experimental Details}  \label{sec:details}

\begin{table}[t]
    \centering
    \small
    \tabcolsep 4pt
    \begin{tabular}{lp{166pt}}
    \toprule
    \textbf{\yn{Date}} & \textbf{ISO-639 Language Codes} \\
    \midrule
    20181001& ru, el, uk, la\\
    \midrule
    \multirow{3}{*}{20181101} & ms, ca, ko, he, fi, ga, ka, th, zh, eu, da, pt, fr, sr, et, sv, hy, cy, sq, hi, hr, bg, ta, sl, bn, id, be, ceb, fa, pl, az, ar, gl, lt, cs, sk, lv, tr, af, vi, ur, ro \\
    \midrule
    20181120& en, nl, ja, it, es, hu, de\\
    \bottomrule
    \end{tabular}
    \caption{\yn{Dates} of downloaded Wikipedia dumps for \yn{the 53 languages supported by mLAMA}.}
    \label{tab:dump_timestamps}
\end{table}

\begin{table*}[t]
    \centering
    \small
    \begin{tabular}{lll}
    \toprule
    \textbf{IDs} & \textbf{Relation} & \textbf{Examples} \\
    \midrule
    \multirow{2}{*}{P103}&  \multirow{2}{*}{The native language of [X] is [Y].} & The native language of Jean-Baptiste Say is French. \\
    & & The native language of Nie Weiping is Chinese. \\
    \midrule
     \multirow{2}{*}{P17}&  \multirow{2}{*}{[X] is located in [Y].} & Noyon is located in France.\\
     & & Gavrilovo-Posadsky District is located in Russia.  \\
     \midrule
     \multirow{2}{*}{P140}&  \multirow{2}{*}{[X] is affiliated with the [Y] religion.} & Abdullah Ahmad Badawi is affiliated with the Islam religion.\\
     & & Noriyasu Hirata is Japan citizen.\\
     \midrule
     \multirow{2}{*}{P1412}&  \multirow{2}{*}{[X] used to communicate in [Y].} & Pere Gimferrer used to communicate in Spanish.\\
     & & Susan McClary used to communicate in English.\\
     \midrule
     \multirow{2}{*}{P27}&  \multirow{2}{*}{[X] is [Y] citizen.} & Priyanka  Vadra is India citizen.\\
     & & Giovanni Lista is Italy citizen.\\
    \bottomrule
    \end{tabular}
    \caption{Relations containing mostly name, country, and location entities.}
    \label{tab:naming_rels}
\end{table*}

\begin{table*}[t]
\centering
\small
\begin{tabular}{cll}
\toprule
\textbf{ISO} & \textbf{Language} & \textbf{\yn{Examples of absent yet predictable facts}} \\
\midrule
af & Afrikaans & Vlag van Jamaika is 'n wettige term in Jamaika. \\
az & Azerbaijani & Split hava limanı Split adını daşıyır. \\
be & Belarusian & \textcyrillic{Сталіцай камуна Гётэбарг з'яўляецца Гётэбарг.}\\
bg & Bulgarian & \textcyrillic{Декларация за създаване на държавата Израел е легален термин в Израел.} \\
ca & Catalan & Govern de Macau és un terme legal en Macau. \\
ceb & Cebuano & Ang Nokia X gihimo ni Nokia. \\
cs & Czech & Guvernér Kalifornie je právní termín v Kalifornie. \\
cy & Welsh & Mae seicoleg cymdeithasol yn rhan o seicoleg. \\
da & Danish & Danmarks Justitsminister er en juridisk betegnelse i Danmark. \\
de & German & Die Hauptstadt von Gouvernorat Bagdad ist Bagdad. \\
el & Greek & \lgrfont{Υπουργός Δικαιοσύνης της Δανίας είναι ένας νομικός όρος στο Δανία.} \\
en & English & Sega Sports R\&D is owned by Sega . \\
es & Spanish & Honda Express es producido por Honda. \\
et & Estonian & Seim (Poola) on Poola -is juriidiline termin. \\
eu & Basque & orbita ekliptiko orbita azpi-klasea da. \\
fi & Finnish & 1955 Dodge tuottaa Dodge. \\
fr & French & Massacre de Cologne se trouve dans Cologne. \\
ga & Irish & Tá Contae Utah suite i Utah. \\
gl & Galician & Sheffield United F.C. recibe o nome de Sheffield. \\
hr & Croatian & Sjedište Valencia C.F. B je u Valencia. \\
hu & Hungarian & Honda Fit -et Honda állítja elő. \\
id & Indonesian & Menteri Kehakiman Denmark adalah istilah hukum dalam Denmark. \\
it & Italian & Nagoya Railroad Co., Ltd è stata fondata a Nagoya. \\
ja & Japanese & \begin{CJK}{UTF8}{ipxm}アンフィオン級潜水艦は潜水艦のサブクラスです。\end{CJK}\\
ko & Korean & \begin{CJK}{UTF8}{mj}모빌군의 수도는 모빌입니다.\end{CJK} \\
la & Latin & Ethica adhibita est pars ethica. \\
lt & Lithuanian & Stokholmas savivaldybė sostinė yra Stokholmas. \\
lv & Latvian & Voterfordas grāfiste galvaspilsēta ir Voterforda. \\
ms & Malay & Sony Alpha 99 dihasilkan oleh Sony. \\
nl & Dutch & Aluminiumsulfaat bestaat uit aluminium. \\
pl & Polish & Cadillac Series 60 jest wytwarzany przez Cadillac. \\
pt & Portuguese & cooperativa autogestionária é uma subclasse de cooperativa. \\
ro & Romanian & Festivalul Internațional de Film de la Calgary este localizat în Calgary. \\
ru & Russian & \textcyrillic{Сенат Теннесси является юридическим термином в Теннесси.} \\
sk & Slovak & BMW N52 sa vyrába v BMW. \\
sl & Slovenian & Narodno gledališče München se nahaja v München. \\
sq & Albanian & BBC Music është pjesë e BBC. \\
sr & Serbian & \textcyrillic{Аеродром Минск је назван по Минск.} \\
sv & Swedish & Huvudstaden till Guvernementet Bagdad är Bagdad. \\
th & Thai & \foreignlanguage{thaicjk}{เมืองหลวงของ เมืองหลวงคอร์ก คือ คอร์ก} \\
tr & Turkish & Waterford County 'un başkenti Waterford' dir. \\
uk & Ukrainian & \textcyrillic{Законодавча асамблея штату Орегон - юридичний термін в Орегон.} \\
vi & Vietnamese & \textviet{Vốn của Hạt Waterford là Waterford.} \\
zh & Chinese & \begin{CJK}{UTF8}{gbsn}意大利\begin{CJK}{UTF8}{bsmi}雜\end{CJK}菜\begin{CJK}{UTF8}{bsmi}湯\end{CJK}是汤的子类。\end{CJK} \\
\bottomrule
\end{tabular}
\caption{Examples of easy-to-predict facts by using shared entity tokens in mLAMA\@.}
\label{table:aaa}
\end{table*}

\subsection{Wikipedia dumps}\label{sec:wikidumps}
\cshin{The Wikipedia data \nc{used} in this study for \nc{assessing the effect of training data volume on factual knowledge acquisition} (\S~\ref{subsec:data_accu}) and tracing the \yn{roots} of facts (\S~\ref{subsec:tracingfacts}) were \nc{taken} from the Wikipedia dumps \nc{in the} Internet Archive.\footnote{The Internet Archive (\url{https://archive.org/}) is a non-profit library of millions of free books, movies, software, music, websites, and more.} We extracted articles in the Main and Article namepspace.\footnote{\url{https://en.m.wikipedia.org/wiki/Wikipedia:What_is_an_article\%3F\#Namespace}}}

\cshin{We collected public Wikipedia dumps for 53 languages, spanning the period between \nc{October 1 and November 20, 2018.}
We chose this timeframe to align 
with mBERT's release date, ensuring \nc{that} data source 
resembled the 
training data of mBERT\@.}
\cshin{The download URLs for each language follow this format: \nolinkurl{https://archive.org/download/{language\_code}wiki-{timestamp}};
\nc{the dumps were downloaded on the basis of data availability during the target period.} The \yn{datas of the downloaded dumps} for each language are listed in Table~\ref{tab:dump_timestamps}.}

\subsection{Rules \nc{for} classifying \yn{types of predictable facts}}\label{sec:factcheck}
We classify the three types of 
predictable facts
\cshin{by}
the following rules. 
\begin{description}
\item[Shared entity tokens:]
We normalized entities by lowercasing strings and unifying Chinese traditional/simplified characters, and then assessed if the object is a substring of or shares subwords with the subject.
Examples of this type can be found in Table~\ref{table:aaa}.
\item[Naming cues:] We manually selected several relations that contain information among person name, location, and countries entities, as illustrated in Table~\ref{tab:naming_rels}. Examples of this type
can be found in Table~\ref{table:bbb}.
\item[Others:] The facts other than those classified into shared tokens across entities and naming cues are regarded as others. Examples of this type
can be found in Table~\ref{table:ccc}.
\end{description}

\begin{table*}[t]
\centering
\small
\begin{tabular}{cll}
\toprule
\textbf{ISO} & \textbf{Language} & \textbf{\yn{Examples of absent yet predictable facts}} \\
\midrule
af & Afrikaans & Die moedertaal van Jean-Baptiste Say is Frans.\\
bg & Bulgarian & \textcyrillic{Родният език на Лионел Жоспен е френски език.}\\
ca & Catalan & La llengua nativa de Alain Mabanckou és francès.\\
ceb & Cebuano & Ang Giovanni Lista usa ka lungsuranon sa Italya.\\
cs & Czech & Rodný jazyk Danielle Darrieuxová je francouzština.\\
cy & Welsh & Mae Guillaumes wedi'i leoli yn Ffrainc.\\
da & Danish & Mødesproget til Pierre Blanchar er fransk.\\
de & German & Die Muttersprache von Pierre Blanchar ist Französisch.\\
en & English & The native language of Hamidou Benmassoud is French .\\
es & Spanish & Bruno Racine solía comunicarse en francés.\\
et & Estonian & Dominic Seiterle on Kanada kodanik.\\
eu & Basque & Umar II.a Islam erlijioarekin erlazionatuta dago.\\
fr & French & Bayazid Bastami est affilié à la religion islam.\\
gl & Galician & Toulouges está situado en Francia.\\
hr & Croatian & Izvorni jezik Jean-Baptiste Say je francuski jezik.\\
hu & Hungarian & John Hutton az angol nyelven történő kommunikációhoz használt.\\
id & Indonesian & Adrian Knox adalah warga negara Australia.\\
it & Italian & La lingua madre di Victor Riqueti de Mirabeau è francese.\\
ja & Japanese & \begin{CJK}{UTF8}{ipxm}ウィリアム・ハウイットの母国語は英語です。\end{CJK}\\
ko & Korean & \begin{CJK}{UTF8}{mj}알랭 마방쿠의 모국어는 프랑스어입니다.\end{CJK}\\
lt & Lithuanian & Gimtoji kalba Nikolajus Dobroliubovas yra rusų kalba.\\
ms & Malay & Bahasa ibunda Jean-Baptiste Say ialah Bahasa Perancis.\\
nl & Dutch & De moedertaal van Jacques Legras is Frans.\\
pl & Polish & Abdolkarim Soroush jest powiązany z religią islam. \\
pt & Portuguese & O idioma nativo de Georges Hugnet é francês.\\
ro & Romanian & Abdolkarim Soroush este afiliat cu religia islam.\\
ru & Russian & \textcyrillic{Насир уд-Дин Абу-л-Фатх Мухаммад связан с религией ислам.}\\
sk & Slovak & Rodný jazyk Vergílius je latinčina.\\
sl & Slovenian & Ernesto Tornquist je državljan Argentina.\\
sq & Albanian & Gjuha amtare e Andrew Jackson është anglisht.\\
sr & Serbian & \textcyrillic{Изворни језик Жан Батист Сеј је француски језик.} \\
sv & Swedish & Sibirkhanatet är anslutet till islam -religionen.\\
tr & Turkish & Guillaumes, Fransa 'da bulunur.\\
vi & Vietnamese & \textviet{Uzhhorod và Moskva là hai thành phố sinh đôi.} \\
zh & Chinese & \begin{CJK}{UTF8}{gbsn}\begin{CJK}{UTF8}{ipxm}円\end{CJK}\begin{CJK}{UTF8}{bsmi}珍,隸屬於佛教宗教\end{CJK}。\end{CJK} \\
\bottomrule
\end{tabular}
\caption{Examples of easy-to-predict facts by using naming cues in mLAMA\@.}
\label{table:bbb}
\end{table*}

\begin{table*}[t]
\centering
\small
\begin{tabular}{cll}
\toprule
\textbf{ISO} & \textbf{Language} & \textbf{\yn{Examples of absent yet predictable facts}} \\
\midrule
af & Afrikaans & Die hoofstad van Verenigde Koninkryk is Londen. \\
az & Azerbaijani & Slovakiya Sosialist Respublikası -nin paytaxtı Bratislava. \\
be & Belarusian & \textcyrillic{Сталіцай Татарская АССР з'яўляецца Казань.} \\
bg & Bulgarian & \textcyrillic{Ембриология е част от медицина.} \\
ca & Catalan & Jean-Baptiste-Claude Chatelain va néixer a París. \\
ceb & Cebuano & Kuala Lumpur (estado) mao ang kapital sa Malaysia. \\
cs & Czech & Beijing College Student Film Festival se nachází v Peking. \\
cy & Welsh & Mae Meade Lux Lewis yn chwarae piano. \\
da & Danish & Jean-Baptiste-Claude Chatelain blev født i Paris. \\
de & German & Surinder Khanna wurde in Delhi geboren. \\
el & Greek & \lgrfont{Πιέρ Λεκόμτ ντου Νουί γεννήθηκε στο Παρίσι.} \\
en & English & Aleksandar Novaković was born in Belgrade . \\
es & Spanish & Aleksandar Novaković nació en Belgrado. \\
et & Estonian & Serbia kuningriik pealinn on Belgrad. \\
eu & Basque & Libano Mendiko eskualdea hiriburua Beirut da. \\
fi & Finnish & Art Davis soittaa jazz -musiikkia. \\
fr & French & Rhigos est un village. \\
ga & Irish & Is é Toulouse príomhchathair Haute-Garonne. \\
gl & Galician & Giuliano Giannichedda xoga na posición centrocampista. \\
hr & Croatian & Glavni grad Narodna Socijalistička Republika Albanija je Tirana. \\
hu & Hungarian & State University of New York székhelye Albany -ben található. \\
id & Indonesian & Ibukota Republik Rakyat Sosialis Albania adalah Tirana. \\
it & Italian & Vernon Carroll Porter è nato a Cleveland. \\
ko & Korean & \begin{CJK}{UTF8}{mj}머피 브라운는 원래 CBS에 방영되었습니다.\end{CJK}\\
la & Latin & Gulielmus Marx Est politicus per professionis. \\
lt & Lithuanian & Ernst \& Young būstinė yra Londonas. \\
lv & Latvian & Itālijas futbola izlase ir loceklis no FIFA. \\
ms & Malay & Power Rangers Samurai pada mulanya ditayangkan pada Nickelodeon. \\
nl & Dutch & Power Rangers: Samurai werd oorspronkelijk uitgezonden op Nickelodeon. \\
pl & Polish & Gregg Edelman to aktor z zawodu. \\
pt & Portuguese & Jean-Baptiste-Claude Chatelain nasceu em Paris. \\
ro & Romanian & Capitala lui Republica Populară Socialistă Albania este Tirana. \\
ru & Russian & \textcyrillic{Штаб-квартира Jim Beam находится в Чикаго.} \\
sk & Slovak & Leicestershire zdieľa hranicu s Lincolnshire. \\
sl & Slovenian & Dilawar Hussain se je rodil v Lahore. \\
sq & Albanian & Guy Doleman është një aktor me profesion. \\
sr & Serbian & \textcyrillic{Сједиште компаније Чикашка берза је у Чикаго.} \\
sv & Swedish & Jean-Baptiste-Claude Chatelain föddes i Paris. \\
tr & Turkish & Aruba Futbol Federasyonu, FIFA üyesidir. \\
uk & Ukrainian & \textcyrillic{Штаб-квартира Партія «Новий Азербайджан» знаходиться в Баку.} \\
vi & Vietnamese & \textviet{Chiếc giày vàng Giải bóng đá Ngoại hạng Anh là một giải thưởng.}\\
zh & Chinese & \begin{CJK}{UTF8}{gbsn}\begin{CJK}{UTF8}{bsmi}拉\end{CJK}尔\begin{CJK}{UTF8}{bsmi}克\end{CJK}·\begin{CJK}{UTF8}{bsmi}沃里斯是\end{CJK}专业\begin{CJK}{UTF8}{bsmi}上的演員\end{CJK}。\end{CJK} \\
\bottomrule
\end{tabular}
\caption{Examples of non-easy-to-predict facts in mLAMA\@.}
\label{table:ccc}
\end{table*}

\end{document}